\documentclass[sigconf,nonacm]{acmart}
\AtBeginDocument{%
  }

\usepackage{amsmath}
\usepackage{hyperref}
\usepackage[nameinlink]{cleveref}
\usepackage{multirow}
\usepackage{arydshln}
\usepackage{subcaption}
\usepackage{lipsum}
\usepackage{xcolor}
\usepackage{CJKutf8}  
\usepackage{booktabs}
\usepackage{amsfonts}
\usepackage{pifont}
\usepackage{fdsymbol}

\usepackage{algorithm}
\usepackage{algpseudocode}
\usepackage{float}
\usepackage{cleveref}
\newcommand{\cmark}{\ding{51}}%
\newcommand{\xmark}{\ding{55}}%

\newcommand{\framework}{\textsc{ConFit v3}}
\newcommand{\confitold}{\textsc{ConFit v2}}
\newcommand{\confitvone}{\textsc{ConFit}}
\newcommand{\confitsimple}{\textsc{ConFit}}
\newcommand{\second}[1]{\textcolor{gray}{\underline{#1}}}

\newcommand{\chinese}[1]{\begin{CJK}{UTF8}{gbsn}#1\end{CJK}}

\begin{document}

\title[\framework{}: Improving Resume-Job Matching with LLM-based Re-Ranking]{\framework{}: Improving Resume-Job Matching with \\ LLM-based Re-Ranking}

\author{Xiao Yu}
\authornote{Equal contribution.}
\affiliation{%
  \institution{Columbia University}
  \city{New York}
  \state{NY}
  \country{USA}
}
\email{xy2437@columbia.edu}

\author{Ruize Xu}
\authornotemark[1]
\affiliation{%
  \institution{Columbia University}
  \city{New York}
  \state{NY}
  \country{USA}
}
\email{rx2246@columbia.edu}

\author{Chengyuan Xue}
\authornotemark[1]
\affiliation{%
  \institution{John Hopkins University}
  \city{Baltimore}
  \state{MD}
  \country{USA}
}
\email{cxue14@jh.edu}

\author{Junyu Chen}
\affiliation{%
  \institution{UCLA}
  \city{Los Angeles}
  \state{CA}
  \country{USA}
}
\email{jochen030327@g.ucla.edu}

\author{Matthew So}
\affiliation{%
  \institution{Columbia University}
  \city{New York}
  \country{USA}
}
\email{ms5513@columbia.edu}

\author{Shijun Ma}
\affiliation{%
  \institution{Intellipro Group Inc.}
  \city{Beijing}
  \country{China}}
\email{ken@intelliprogroup.com}

\author{Bo Liu}
\affiliation{%
  \institution{Intellipro Group Inc.}
  \city{Beijing}
  \country{China}}
\email{bo.liu@altomni.com}

\author{Xiangye Liang}
\affiliation{%
  \institution{Intellipro Group Inc.}
  \city{Beijing}
  \country{China}}
\email{galabala.liang@intelliprogroup.com}

\author{Zhou Yu}
\affiliation{%
  \institution{Columbia University}
  \city{New York}
  \country{USA}
}
\email{zy2461@columbia.edu}
\renewcommand{\shortauthors}{Yu et al.}

\begin{abstract}
  A reliable resume-job matching system helps a company find suitable candidates from a pool of resumes and helps a job seeker find relevant jobs from a list of job posts.
  While recent advances in embedding-based methods such as \confitvone{} and \confitold{} can efficiently retrieve candidates at scale, the lack of controllability and explainability limits their real-world adaptations.
  LLM-based re-rankers can address these limitations through reasoning, but existing training recipes are developed on short-document benchmarks and do not account for noise in real-world recruiting data.
  In this work, we first conduct a systematic analysis over the LLM re-ranker training pipeline for person-job fit, covering inference algorithm design, RL algorithm selection, data processing, and SFT distillation.
  We find that using multi-pass re-ranking, training with listwise RL objectives, removing noisy samples, and distilling from a stronger LLM before RL significantly improves re-ranking performance.
  We then aggregate these findings to train \framework{} with Qwen3-8B and Qwen3-32B on real-world person-job fit datasets, and find significant improvements over existing best person-job fit systems as well as strong LLMs such as GPT-5 and Claude Opus-4.5.
  We hope our findings provide useful insights for future research on adapting LLM-based re-rankers to person-job fit systems\footnote{Code available at \url{https://github.com/jasonyux/ConFit-v3.git}}.
\end{abstract}




\maketitle

\section{Introduction}

Online recruitment platforms such as LinkedIn serve over one billion users with diverse and complex hiring needs \citep{linkedin}, requiring recommendation systems that can accurately match resumes and jobs at scale.
Embedding-based methods \citep{E5,zhang2025qwen3embeddingadvancingtext,vera2025embeddinggemmapowerfullightweighttext}, such as \confitvone{} \citep{confit_v1} and \confitold{} \citep{yu2025confitv2improvingresumejob}, have shown strong results for person-job fit by efficiently matching resumes and jobs in a shared embedding space.
However, these methods lack controllability and explainability: they offer limited ability to prioritize results based on domain-specific criteria, and provide no reasoning for why one candidate is ranked above another.

\begin{figure}[t!]
    \centering
    \includegraphics[scale=0.81]{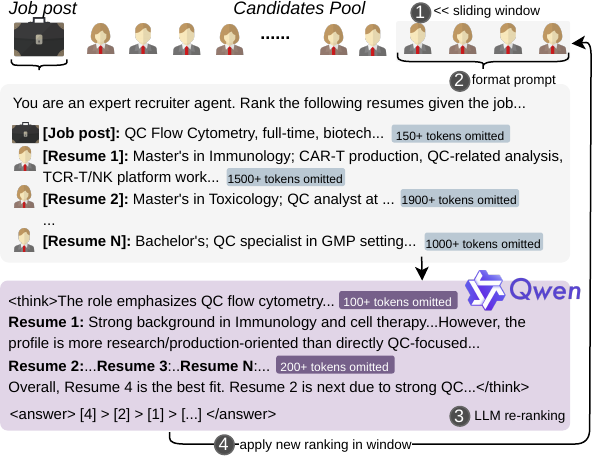}
    \caption{Overview of LLM-based re-ranking in \framework{}. Given a job post and a pool of candidate resumes (e.g., ranked by \confitold{}), we train Qwen3-8B/Qwen3-32B to iteratively re-rank the resumes using a sliding-window based method.}
    \Description{Overview of LLM-based re-ranking in \framework{}. Given a job post and a pool of candidate resumes (e.g., ranked by \confitold{}), we train Qwen3-8B/Qwen3-32B to iteratively re-rank the resumes using a sliding-window based method.}
    \label{fig:overall_fig}
\end{figure}

To address these limitations, recent work has turned to large language models (LLMs) for re-ranking, leveraging their strong reasoning capabilities \citep{brown2020languagemodelsfewshotlearners,ahn2024largelanguagemodelsmathematical,jimenez2024swebenchlanguagemodelsresolve,wang2024largelanguagemodelsrobotics} to compare and re-order passages with more accurate and interpretable rankings \citep{liang2023holisticevaluationlanguagemodels,rankgpt,qin2024largelanguagemodelseffective}.
However, existing recipes that finetune LLMs as re-rankers \citep{pradeep2023rankvicunazeroshotlistwisedocument,pradeep2023rankzephyreffectiverobustzeroshot,zhuang2025rankr1enhancingreasoningllmbased,zhang2025rearankreasoningrerankingagent} are primarily developed on academic benchmarks such as MTEB \citep{muennighoff2022mteb,enevoldsen2025mmtebmassivemultilingualtext}, where documents are short and labels contain little noise.
Person-job fit presents two distinct challenges: (1) resumes and job posts are long documents (often exceeding thousands of tokens); and (2) recruiting data is inherently noisy, as many highly competitive candidates are unlabelled simply because they never applied, not because they were unqualified.

In this work, we systematically study modern LLM-based re-ranker training recipes in the context of person-job fit, analyzing which design choices most affect performance.
Our empirical study reveals three key findings.
First, multi-pass re-ranking at inference time substantially improves performance over the single-pass approach used by most prior work \citep{rankgpt,pradeep2023rankvicunazeroshotlistwisedocument,pradeep2023rankzephyreffectiverobustzeroshot}.
Second, label quality matters more than quantity during RL training. Removing noisy samples is more effective than sophisticated filtering strategies.
Third, SFT distillation from a stronger model significantly improves capability and provides a better initialization for RL.

Building on these findings, we train \framework{} by aggregating our findings and fine-tuning Qwen3-8B/Qwen3-32B \citep{yang2025qwen3technicalreport}.
We evaluate on two real-world person-job fit datasets and find significant improvements over prior best retrieval-based systems as well as strong LLM-based re-rankers such as GPT-5 \citep{singh2025openaigpt5card} and Claude Opus-4.5 \citep{opus45}.
We hope our empirical analysis and results provide useful insights for future research on adapting LLM-based re-rankers to person-job fit systems.

\section{Background}
\label{sec:Background}
A resume-job matching (or \emph{person-job fit}) system models the compatibility between a resume $R$ and a job post $J$ to recommend suitable candidates for a given job or relevant jobs for a given candidate \cite{mvcon,DPGNN,InEXIT,confit_v1,yu2025confitv2improvingresumejob}.
While embedding-based approaches \citep{confit_v1,yu2025confitv2improvingresumejob} can efficiently rank tens of thousands of passages at scale, they lack controllability and explainability as the ranking process happens entirely in the latent embedding space.

To address this, recent work \citep{rankgpt,pradeep2023rankvicunazeroshotlistwisedocument,zhang2025rearankreasoningrerankingagent} explores using the strong reasoning ability of LLMs to re-rank passages.
We briefly illustrate this in \Cref{fig:overall_fig}.
Since LLMs are more costly than embedding-based approaches, in practice they are often used for \emph{re-ranking} top-$N$ passages retrieved by embedding-based approaches, where $N$ is typically small.
Formally, given a job post $J$ and an initial ordered list of candidate resumes $\mathcal{R} = (R_1, R_2, \ldots, R_N)$ retrieved by an embedding model, the LLM re-ranker produces a re-ordered list:
\[
(R_{\pi(1)}, R_{\pi(2)}, \ldots, R_{\pi(N)}) = \textrm{LLM}(J, \mathcal{R}),
\]
where $\pi$ is a permutation over $\{1, \ldots, N\}$ ranking the most relevant resumes first (ideally the resume that is accepted by the job).
In this work, we consider the re-ranking task as re-ordering the top-20 resumes for a given job post, where resumes average 1237.7 tokens and job posts average 575.6 tokens.
Additionally, since a candidate usually only applies to a few positions, interaction labels $y_{i} \in \{0, 1\}$ representing whether a resume $R_i$ is accepted by a job $J$ are very sparse and often noisy/subjective.
For example, in both the Recruiting and AliYun datasets, less than 0.05\% of the total possible (resume, job) pairs are annotated (see \Cref{appendix:More Details on Dataset and Preprocessing} for more details).
We note that this setting is in sharp contrast to many benchmarks such as MTEB used by recent work \citep{zhuang2025rankr1enhancingreasoningllmbased,zhang2025rearankreasoningrerankingagent}, where passages are typically only a few \emph{hundred} tokens and labels contain little noise.

\section{LLM Re-Ranker Analysis}
\label{sec:re-ranker training analysis}
This section explores and quantifies which choices are important to train LLM-based re-ranker for person-job fit. We introduce our experimental setup in \Cref{subsec:analysis_exp_setup}, and then study the impact of various design choices in \Cref{subsec:Re-Ranking Inference Algorithms,subsec:RL Algorithms for Re-ranking,subsec:RL Data Processing,subsec:SFT Distillation}.

\subsection{Experimental Setup}
\label{subsec:analysis_exp_setup}

\paragraph{Dataset and Models} For a controlled study, we perform our study on the Recruiting dataset and Qwen3-8B \citep{zhang2025qwen3embeddingadvancingtext} model (in \Cref{sec:ConFitv3} we extend to other datasets and models).
The Recruiting dataset is a proprietary dataset provided by a hiring solution company, containing anonymized resume-job pairs with accept/reject labels.
To convert the dataset into a re-ranking dataset, we use \confitold{} to retrieve the top-$N$ resumes per job and construct training windows from the ranked list. See \Cref{appendix:More Details on Dataset and Preprocessing,appendix:More Details on Dataset Postprocessing} for more details on dataset statistics and training data construction, respectively.

\paragraph{Evaluation Metrics} Given a job post, we evaluate the LLM's ability to \emph{re-rank} top-20 resumes retrieved by a finetuned embedding model.
Specifically, we use \confitold{} as the embedding model as it achieved the prior best result on similar person-job fit datasets \citep{yu2025confitv2improvingresumejob}.
We use standard metrics such as nDCG@10 and Recall@10 to evaluate the LLM's ability to identify the top-10 most suitable resumes amongst the retrieved pool of 20 highly competitive resumes.

\subsection{Inference Algorithms for Re-Ranking}
\label{subsec:Re-Ranking Inference Algorithms}
Given a powerful LLM, we first study how inference-time algorithms and configurations affect re-ranking performance.
Since resumes and job posts are often long, re-ranking the top-20 in a single pass would produce prohibitively long contexts for both training and inference.
We therefore adopt the sliding-window method introduced by \citet{rankgpt,pradeep2023rankvicunazeroshotlistwisedocument,zhang2025rearankreasoningrerankingagent}, and study how hyperparameters such as window size, stride, and number of iterations affect performance.
Specifically, given a job post and a list of $N$ long resumes, the inference algorithm 1) prompts the LLM to rank a window of $k << N$ resumes, often starting at the bottom of the list; 2) shifts the window by $s$ resumes, and 3) repeats this process until the window reaches the top of the list.
Finally, since running this process once often cannot significantly change the ordering of the resumes, the entire algorithm can be repeated multiple times $t$. Please see \Cref{sec:details_on_inference_algorithm} for pseudocode.

We experimented with a combination of different $k$, $s$, and $t$, and found that changing the number of iterations $t$ is the most impactful hyperparameter.
In \Cref{tbl:scaling_results}, we find that (1) increasing $t=1$ to $t=2$ significantly improves performance; and (2) further increasing $t$ shows diminishing returns.
This suggests that while the LLM can rank well within each window, a single pass (used by most work) is often insufficient for propagating the best candidates to the top of the list.
Two passes allow the algorithm to refine the ordering and largely converge, while further iterations yield diminishing returns, suggesting the LLM's ranking ability is already well-utilized after two iterations.
For a fair comparison, we will use $t=2$ for the rest of our analysis in \Cref{sec:re-ranker training analysis}, unless otherwise specified.
For results on other hyperparameters such as $k$ and $s$, please see \Cref{sec:details_on_inference_algorithm}.

\begin{table}[t]
  \caption{Iterative re-ranking performance}
  \label{tbl:scaling_results}
  \centering
  \begin{tabular}{l l c c c}
  \toprule
   & $t=1$ & $t=2$ & $t=3$ & $t=4$ \\
  \midrule
  nDCG@10 & 55.41 & 57.96 & 57.83 & 57.94\\
  Recall@10 & 64.74 & 67.10 & 69.03 & 67.27 \\
  Average & 60.08 & 62.53 & 63.43 & 62.61 \\
  \bottomrule
  \end{tabular}
\end{table}
\begin{table}[t]
    \centering
    \caption{Comparison of RL algorithms for re-ranking.}
    \label{tbl:rl_algo_cmp_results}
    \begin{tabular}{l c c c}
    \toprule
    Method & nDCG@10 & Recall@10 & Average \\
    \midrule
     -- (Prompt) & 47.52   & 60.37 & 53.95\\
    \midrule
    Rank-R1 & 52.36 & 62.79 & 57.58 \\
    ReaRank & \textbf{55.00} & \textbf{65.60} & \textbf{60.30} \\
    \bottomrule
    \end{tabular}
\end{table}

\subsection{RL Algorithms for Re-ranking}
\label{subsec:RL Algorithms for Re-ranking}
Optimizing an LLM-based re-ranker beyond prompting often involves fine-tuning the LLM on relevant datasets.
Following recent advances in reinforcement learning (RL), we investigate two recipes for training LLM-based re-rankers via RL.
(1) Rank-R1 \cite{zhuang2025rankr1enhancingreasoningllmbased} which largely treats ranking as a ``classification'' task and prompts the LLM to locate the top-1 passage within each window, optimized using a binary 0-1 reward based on whether the LLM correctly identifies the positive passage in the window.
(2) ReaRank \cite{zhang2025rearankreasoningrerankingagent} which prompts the LLM to output a complete ordering list of the passages in the window (e.g., see \Cref{fig:overall_fig}), optimized using relative nDCG improvement as reward:
\begin{equation}
\label{rearank_formula}
 R=\frac{\mathrm{nDCG}_{\text{new}}-\mathrm{nDCG}_{\text{old}}}{\mathrm{nDCG}_{\text{max}}-\mathrm{nDCG}_{\text{old}}},   
\end{equation}
where \(\mathrm{nDCG}_{\text{old}}\), \(\mathrm{nDCG}_{\text{new}}\), and \(\mathrm{nDCG}_{\text{max}}\) denote the original, re-ranked, and maximum nDCG, respectively. For our settings where only one positive candidate exists in the window, $\mathrm{nDCG}_{\text{max}}\equiv1$.
Both methods are then optimized using Group Relative Policy Optimization (GRPO) \cite{shao2024deepseekmathpushinglimitsmathematical,deepseekr1}.
For more details, please see \Cref{appendix:details on RL implementation}.

We present the results in \Cref{tbl:rl_algo_cmp_results}. In \Cref{tbl:rl_algo_cmp_results}, we find ReaRank outperforms Rank-R1 by a significant margin across both nDCG@10 and Recall@10.
We attribute this to the alignment between training and inference: ReaRank trains the LLM to produce a full list-wise ordering, more closely matching the inference task. In contrast, Rank-R1 only asks the model to identify a single top passage, which provides a weaker learning signal.
Since ReaRank is more effective, we will use it as the main RL method for subsequent analysis.

\begin{table}[t]
    \centering
    \caption{Effect of different data processing methods for RL training (based on ReaRank).}
    \label{tbl:data_filtering_cmp_results}
    \begin{tabular}{l l c c}
    \toprule
    Data Method & $|D_{\text{RL}}|$ & nDCG@10 & Recall@10 \\
    \midrule
    - (All Data) & 11.7k & 55.00 & 65.60 \\
    \midrule
    Subsample Hard & 8.0k & 53.37 & 63.85 \\
    LLM Filter & 5.5k & 56.71 & 66.58 \\
    Hint Augment (\cite{pope}) & 4.4k & 55.62 & 63.65 \\
    Remove Hard (ours) & 4.3k & \textbf{57.96} & \textbf{67.10} \\
    \bottomrule
    \end{tabular}
\end{table}
\subsection{RL Data Processing}
\label{subsec:RL Data Processing}
Beyond the choice of RL algorithm, the quality of training data also plays a critical role.
For example, \citet{yu2025dapoopensourcellmreinforcement} find that filtering out samples that are ``too easy'' ($r=1.0$ for the entire group) or too hard ($r=0.0$ for the entire group) leads to more effective GRPO training.
Motivated by this, we investigate several data processing methods for RL-based re-ranker training.
We consider: (1) \emph{Subsample Hard}: subsampling hard samples (average reward $\bar{r} < 0.4$) by 50\%; (2) \emph{LLM Filter}: prompting an LLM-as-a-judge to select samples suitable for learning; (3) \emph{Hint Augment} \citep{pope}: adding hints for hard samples during RL; and (4) \emph{Remove Hard}: removing all samples with $\bar{r} < 0.4$.

We present results in \Cref{tbl:data_filtering_cmp_results}, and find that simply removing all hard samples (\emph{Remove Hard}) achieves the best performance.
Upon manual inspection, we find that many hard samples contain competitive resumes that appear stronger than the accepted candidates.
These resumes were often unlabelled likely because the candidates may not have applied to that position (e.g., not aware of it or not interested in it), rather than being unqualified.
As a result, these samples introduce noisy supervision that harms RL learning.
We believe relabeling these data by experts could further improve RL training, which we leave for future work.

\begin{table}[t]
    \centering
    \caption{Effect of SFT distillation prior to RL training}
    \label{tbl:sft_comparison}
    \begin{tabular}{l c c c}
    \toprule
    Method & nDCG@10 & Recall@10 & Average \\
    \midrule
    SFT-only & 59.51 & 67.48 & 63.50 \\
    RL-only & 57.96 & 67.10 & 62.53 \\
    SFT+RL & \textbf{60.83} & \textbf{68.40} & \textbf{64.62} \\
    \bottomrule
    \end{tabular}
\end{table}
\subsection{SFT Distillation}
\label{subsec:SFT Distillation}
Lastly, we investigate whether distilling from a stronger model via SFT before RL can further improve performance \cite{yue2025doesreinforcementlearningreally}.
Specifically, we (1) collect demonstration data by prompting Claude Sonnet 4.5\footnote{We use Claude Sonnet 4.5 to better balance cost, quality, and quantity.} to rank resumes in our training set; (2) only keep generations where the accepted resume is correctly placed at the top of the ordering, and (3) SFT Qwen3-8B on the resulting data prior to RL.

We present results in \Cref{tbl:sft_comparison}. We find SFT-only already outperforms RL-only, indicating the challenging nature of person-job fit for smaller LLMs.
Combining SFT with RL (SFT+RL) achieves the best performance, suggesting that distillation provides a better initialization for RL to further refine.
This result is also consistent with many general RL works such as \citep{zhang2025qwen3embeddingadvancingtext, ouyang2022traininglanguagemodelsfollow,yu2025dynamindlearningsimulateexperience,wei2025swerladvancingllmreasoning}.

\begin{table*}[t]
  \centering
  \caption{Comparing \framework{} to multiple training and prompting methods on the Recruiting Dataset.
  $|D_{\text{SFT}}|$ and $|D_{\text{RL}}|$ denotes the training dataset size during SFT and RL, respectively.
  Best result is shown in \textbf{bold}, runner-up is shown in \second{gray}.}
  \label{tab:recruiting_dataset_scores}
  \begin{tabular}{l l c c c c c}
  \toprule
  Re-ranking Method & Training & $|D_{\text{SFT}}|$ & $|D_{\text{RL}}|$ & nDCG@10 & Recall@10 & Average \\
  \midrule

  -- (ConFit v2) & \xmark & -- & -- & 52.33 & 62.30 & 57.32 \\
  \midrule
  \multicolumn{4}{l}{\textbf{Prompting}} \\
  Qwen3-8B & \xmark & -- & -- & 48.97 & 61.04 & 55.01 \\
  Qwen3-32B & \xmark & -- & -- & 54.10 & 63.57 & 58.84 \\
  Qwen3-235B-A22B & \xmark & -- & -- & 55.43 & 64.30 & 59.87 \\
  GPT-4.1 & \xmark & -- & -- & 57.72 & 65.59 & 61.66 \\
  Claude Opus-4.5 & \xmark & -- & -- & 57.87 & 65.42 & 61.65 \\
  GPT-5 & \xmark & -- & -- & 57.43 & 65.32 & 61.38 \\

  \midrule
  \multicolumn{6}{l}{\textbf{SFT (Qwen3-8B)}} \\
  Distill (Sonnet-4.5) & \cmark & 4.4k & -- & 57.67 & 64.98 & 61.33 \\
  \midrule
  \multicolumn{6}{l}{\textbf{RL (Qwen3-8B)}} \\
  Rank-R1 & \cmark & -- & 11.7k & 52.14& 62.48 & 57.31 \\
  ReaRank & \cmark & -- & 11.7k & 52.44 & 62.30 & 57.37 \\
  \midrule
  \framework{} (Qwen3-8B) & \cmark & 4.4k & 4.3k & \second{60.83} & \second{68.40} & \second{64.62} \\
  \framework{} (Qwen3-32B) & \cmark & 4.4k & 5.3k & \textbf{61.37} & \textbf{68.89} & \textbf{65.13} \\
  \bottomrule
  \end{tabular}
\end{table*}
\begin{table}[t]
  \centering
  \caption{Results on the AliTianChi Dataset. All training methods uses Qwen3-8B. *Since AliTianChi is easier, we omit the SFT stage prior to RL for \framework{}.}
  \label{tab:aliyun_results}
  \begin{tabular}{l  c c c}
  \toprule
  
  Method & $|D_{\text{RL}}|$ & nDCG@10 & Recall@10 \\
  \midrule 
  -- (ConFit v2) & - & 80.28 & 95.72 \\
  \midrule
  Prompt (Qwen3-8B) & - & 72.43 & 95.77 \\
  Prompt (Qwen3-32B) & - & 74.20 & 95.38 \\
  Prompt (GPT-5) & - & 74.81 & 96.02\\
  ReaRank & 13.3k & 77.47& 95.46\\
  \midrule
  \framework{}* & 5.7k & \textbf{82.65} & \textbf{96.91} \\
  \bottomrule
  \end{tabular}
\end{table}

\section{\framework{}}
\label{sec:ConFitv3}
The previous sections empirically identified effective strategies for training and running an LLM-based re-ranker for person-job fit.
We now aggregate these findings to build \framework{} and evaluate it on two real-world datasets.

\subsection{Experimental Setup}
\label{subsec:ConFitv3 Experimental Setup}

\paragraph{Training and Inference Recipe} We aggregate our findings in \Cref{sec:re-ranker training analysis} into a single recipe for LLM-based re-ranker training and inference. During training, we (1) first distill from a stronger model via SFT; (2) then use ``Remove Hard'' to post-process training data for RL; and (3) use ReaRank to perform RL training, continuing from the SFT checkpoint. During inference, we use the sliding-window based method but with $t=2$ iterations.

\paragraph{Datasets and Models} Following \Cref{sec:re-ranker training analysis}, we first consider applying our recipe to Qwen3-8B on the proprietary Recruiting Dataset. However, to ensure the generalizability of our findings, we further apply our recipe to (1) training Qwen3-32B on the Recruiting Dataset; and (2) training Qwen3-8B on the AliYun Dataset, a dataset from the 2019 Alibaba job-resume matching competition containing desensitized resume-job pairs.

\paragraph{Baselines} We first compare against the prior best retrieval-based system \confitold{} \citep{yu2025confitv2improvingresumejob} without using LLM-based re-ranker.
Then, we compare against (1) prompting several strong open- and closed-sourced models such as Qwen3-235B-A22B-Instruct-2507 \citep{yang2025qwen3technicalreport}, GPT-5 \citep{singh2025openaigpt5card}, and Opus 4.5 \citep{opus45} as re-ranker; and (2) applying recent training-based methods such as Rank-R1 \citep{zhuang2025rankr1enhancingreasoningllmbased} and ReaRank \citep{zhang2025rearankreasoningrerankingagent}.

\subsection{Main Results}
\label{subsec:Main Results}
We present results on the Recruiting Dataset in \Cref{tab:recruiting_dataset_scores} and the AliYun Dataset in \Cref{tab:aliyun_results}.
On the Recruiting Dataset, our best result (\framework{} (Qwen3-32B)) improves upon \confitold{} by 7.81\% absolute on average.
Compared to other training-based methods using the same Qwen3-8B backbone, \framework{} also achieves significant gains.
When extending to the AliYun Dataset, in \Cref{tab:aliyun_results} we find our recipe similarly achieves the best performance.
We believe these results demonstrate the effectiveness of our recipe for training LLM-based re-rankers for person-job fit systems. For ablation studies, limitations, and ethical considerations, please see \Cref{appendix:Ablation,appendix:Limitations,appendix:Ethical Considerations}, respectively. For appendices, please see \href{https://anonymous.4open.science/r/confit_v3_supplementary-0B95/sample-sigconf.pdf}{this anonymous github}.

\section{Related Work}
\label{sec:Related Work}

\paragraph{Person-job fit systems}

Early person-job fit systems build on neural architectures such as RNNs, LSTMs, and CNNs \citep{APJFNN,pjfnn,cnn-lstm-siamese,jiang2020learning,10169716,staudemeyer2019understanding,oshea2015introduction}.
These are later outperformed by transformer-based methods \citep{siamese,bian-etal-2019-domain,Zhang2023FedPJFFC,devlin2019bert}.
However, these transformer-based methods often rely on task-specific assumptions that limit generalization, such as requiring all resumes and jobs to be seen during training or requiring access to privileged interaction signals.
More recently, \confitvone{} \citep{yu2024confitimprovingresumejobmatching} and \confitold{} \citep{yu2025confitv2improvingresumejob} show that dense retrieval with contrastive learning and hard-negative mining can significantly improve person-job fit using a general-purpose encoder like E5 \citep{E5}.
However, these embedding-based methods lack controllability and explainability.
To the best of our knowledge, \framework{} is the first to apply LLM-based re-ranking approaches to person-job fit, addressing these limitations while further improving performance.

\paragraph{LLMs for re-ranking} Advances in LLM reasoning have enabled a wide range of applications, including passage re-ranking.
Recent work explores various comparison prompting-based strategies such as pointwise \citep{liang2023holisticevaluationlanguagemodels}, pairwise \citep{qin2024largelanguagemodelseffective}, setwise \citep{Zhuang2024}, and listwise \citep{ma2023zeroshotlistwisedocumentreranking,pradeep2023rankvicunazeroshotlistwisedocument,rankgpt} approaches to perform re-ranking.
Beyond prompting, methods such as Rank-R1 \citep{zhuang2025rankr1enhancingreasoningllmbased}, ReaRank \citep{zhang2025rearankreasoningrerankingagent}, and more \citep{pradeep2023rankvicunazeroshotlistwisedocument,pradeep2023rankzephyreffectiverobustzeroshot} further optimize re-rankers through SFT and RL training.
However, these methods primarily focus on academic benchmarks such as TREC \citep{craswell2020overviewtrec2019deep}, BEIR \citep{thakur2021beir}, and MTEB \citep{muennighoff2022mteb,enevoldsen2025mmtebmassivemultilingualtext}, where passages are typically only a few hundred words and labels contain little noise.
In contrast, person-job fit involves ranking passages (i.e., resumes and job posts) with thousands of tokens and training with inherently noisy labels. Our work presents a systematic analysis of adapting these methods to this setting, and we hope it can serve as a reference point for future research on LLM-based re-ranking for person-job fit.

\section{Conclusion}
\label{sec:Conclusion}
We present a detailed analysis of adapting modern LLM-based re-ranking methods to person-job fit, and find that multi-pass re-ranking, training with listwise RL objectives, removing noisy samples, and distilling from a stronger LLM before RL significantly improves re-ranking performance.
We apply this recipe to train \framework{} and evaluate it on two real-world person-job fit datasets.
We find \framework{} outperforms both prompting state-of-the-art LLMs such as GPT-5 and Claude Opus 4.5, as well as directly adapting existing training-based methods such as Rank-R1 and ReaRank.
These results highlight the importance of addressing challenges specific to person-job fit, and provide a reference for future work on LLM-based re-ranking in this domain.

\bibliographystyle{ACM-Reference-Format}
\bibliography{custom}

\clearpage
\appendix

\setcounter{table}{0}
\renewcommand{\thetable}{A\arabic{table}}
\setcounter{figure}{0}
\renewcommand{\thefigure}{A\arabic{figure}}

\section{Limitations}
\label{appendix:Limitations}

This is due to the highly sensitive nature of resume and job post content, making large-scale person-job fit datasets largely proprietary.
We follow \citet{confit_v1,yu2025confitv2improvingresumejob} and provide the best effort to make \framework{} reproducible and extensible for future work.
We will open-source implementations of \framework{}, related baselines, data processing scripts, and dummy train/valid/test data files that can be used to test drive our system end-to-end.

\section{Ethical Considerations}
\label{appendix:Ethical Considerations}


Since \framework{} relies on fine-tuning pretrained LLMs, it may inherit biases present in both the pretraining data and the recruiting dataset, such as preferences related to gender, race, or demographics \citep{bai2024measuringimplicitbiasexplicitly,an2024measuringgenderracialbiases,gallegos2024biasfairnesslargelanguage}.
To mitigate this, we follow \citep{confit_v1,yu2025confitv2improvingresumejob} and removed all sensitive identity information (e.g., gender, name, age) from the training data during preprocessing.
We do \textbf{not} condone using \framework{} for real-world applications without using de-biasing techniques such as \citep{gaci-etal-2022-debiasing,guo-etal-2022-auto,schick-etal-2021-self,tan2025analyzinggeneralizationreliabilitysteering,siddique2026shiftingperspectivessteeringvectors}, and in general, we do \textbf{not} condone the use of \framework{} for any morally unjust purposes.
To our knowledge, there is little work on investigating or mitigating biases in existing person-job fit systems or LLM-based re-rankers, and we believe this is a crucial direction for future person-job research.

\section{Ablation Study}
\label{appendix:Ablation}

To complement our analysis in \Cref{sec:re-ranker training analysis}, we perform an ablation study detailing the performance contribution for each component of our recipe.
We use Qwen3-8B on the Recruiting Dataset as an example, and present the results in \Cref{tab:ablation_results}.
Specifically, we compare (1) \emph{no RL data prc.} which skips the ``Remove Hard'' step used for RL data post-processing; (2) \emph{no SFT} which skips the SFT stage prior to RL; (3) \emph{no inf. alg.} which uses the default sliding-window based inference algorithm with $t=1$; and (4) \emph{no training} which performs no SFT nor RL training, i.e., directly prompting the LLM to re-rank the resumes.

Overall, we find that removing noisy training labels and using multi-pass sliding-window inference impacts the final performance the most.
This suggests that (1) label quality matters more than quantity for RL training; and (2) proper inference configurations are essential to fully utilize the LLM's ranking capabilities.

\section{More Details on LLM re-ranker Prompts}
\label{appendix:more_detail_on_rerank_prompt}

We present the full prompt template used by \textsc{ConFit v3} in
Table~\ref{tab:prompt_template}. Given a job post $J$ and a window of
$k=4$ candidate resumes $W = \{R_1, R_2, R_3, R_4\}$, the LLM is instructed to (1) summarize the job post
and briefly analyze each resume, (2) compare the resumes against each
other, and (3) produce a full ordering enclosed in
\texttt{<answer>}~tags (e.g., \texttt{<answer> [2] > [4] > [1] > [3] </answer>}).
This listwise format is used consistently for prompting, SFT distillation, and RL training.

To make the prompt \emph{controllable}, we include an explicit
``non-negotiable requirements'' checklist covering education,
certifications, technical skills, age restrictions, legal and identity
requirements, physical fitness, and work-condition constraints. This
checklist is inspired by heuristics used by real recruiters and allows
practitioners to inject domain-specific hiring criteria without
retraining.

\begin{table}[t]
  \centering
  \caption{Ablation study of \framework{} based on Qwen3-8B.}
  \label{tab:ablation_results}
  \begin{tabular}{l c c c c}
  \toprule
  Method & $|D_{\text{SFT}}|$ & $|D_{\text{RL}}|$ & nDCG@10 & Recall@10 \\
  \midrule
  \framework{} & 4.4k & 4.3k & \textbf{60.83} & \textbf{68.40} \\
  \, - no RL data prc. & 4.4k & 11.7k & 57.04 & 63.83 \\
  \, - no SFT & -- & 4.3k & 57.96 & 67.10 \\
  \, - no inf. alg. & -- & 4.3k & 55.41 & 64.74 \\
  \, - no training & -- & -- & 48.97 & 61.04 \\
  \bottomrule
  \end{tabular}
\end{table}
\begin{table}[!t]
  \centering
  \caption{Dataset statistics. ``\# Token per $R$/$J$'' represent the \emph{mean($\pm$std)} number of token per resume/job after post-processing.}
  \label{tbl:train_dset}
  \scalebox{0.99}{
    \begin{tabular}{l cc}
      \toprule
       & \textbf{Recruiting Dataset} & \textbf{Aliyun Dataset}\\
      \midrule
      \# Jobs     & 10597 & 19542 \\
      \# Resumes  & 49398 & 2718 \\
      \# Labels   & 58111 & 22124 \\
      \phantom{--}\textcolor{gray}{(\# accept)} & 27421 & 10185 \\
      \phantom{--}\textcolor{gray}{(\# reject)} & 30690 & 11939 \\
      \cmidrule(lr){1-3}
      \# Token per $R$ & 1237.7($\pm 1012.5$) & 250.6($\pm 96.5$) \\
      \# Token per $J$ & 575.6($\pm 406.8$) & 335.1($\pm 143.9$) \\
      \bottomrule
    \end{tabular}
  }
\end{table}
%
%
\begin{table}[!t]
  \centering
  \caption{Total training and test dataset statistics (prior to the post-processing described in \Cref{subsec:RL Data Processing}) used for re-ranking. ``Train'' samples represent individual windows of 4 resumes, while test samples represent full top-20 re-ranking tasks.
  }
  \label{tbl:test_dset}
  \scalebox{0.99}{
    \begin{tabular}{l ccc ccc}
      \toprule
      & \multicolumn{2}{c}{\textbf{Recruiting Dataset}} & \multicolumn{2}{c}{\textbf{AliYun Dataset}}\\
       & Train & Test & Train & Test \\
      \midrule
      \# Samples
      & 11,657 & 200
      & 13,180 & 300\\
      \# Jobs
      & 6839 & 175
      & 8292 & 289\\
      \# Resumes
      & 23343 & 1142
      & 3713 & 1012\\
      \bottomrule
    \end{tabular}
  }
\end{table}

We further illustrate the prompt with one example from the Recruiting
Dataset in Table~\ref{tab:prompt_example}. The example shows a
job post for an ``Enterprise Account Manager'' role together with four
candidate resumes, one of which (randomly shuffled among the four
positions) is the accepted candidate. Sensitive information such as
names, contact details, and company names has either been removed or
replaced with numeric identifiers.

\begin{table*}[t]
\centering
\small
\caption{Prompt template used by \textsc{ConFit v3} for listwise
re-ranking. Example \texttt{\{additional\_instructions\_or\_heuristics\}}, \texttt{\{job\_description\}} and \texttt{\{resumes\_section\}} can be found in \Cref{tab:prompt_example}.}
\label{tab:prompt_template}
\begin{tabular}{p{0.96\textwidth}}
\toprule
\textbf{System Prompt} \\
\midrule
You are an expert technical recruiter that can rank resumes based on
their matching degree to the job description. You first analyze each
resume individually, then compare them systematically, and finally
provide the ranking. ..(truncated). The most relevant resumes should
be listed first. The output format should be
\texttt{<answer> [] > [] > [] > [] </answer>},
e.g., \texttt{<answer> [X] > [Y] > [Z] > [T] </answer>}. \\
\midrule
\textbf{User Prompt} \\
\midrule
\texttt{\{additional\_instructions\_or\_heuristics\}} \\ 
\textbf{Resumes}: \texttt{\{resumes\_section\}} \\
Please rank these resumes according to their matching degree to the
JOB DESCRIPTION: \texttt{[\{job\_description\}]}. \\
\\
Follow these steps exactly: \\
1. First, think to summarize the job description and analyze EACH resume briefly: Evaluate how well it matches the job description and mandatory criteria.
\\
2. Then, think to COMPARE the resumes and determine which candidates are better fits and why. \\
3. Finally, within <answer> tags, provide ONLY the final ranking of the resumes from best to worst fit using their numerical identifiers in the format: [X] > [Y] > [Z] > [T]. \\

\bottomrule
\end{tabular}
\end{table*}

\begin{table*}[t]
\centering
\small
\caption{One illustrative example of the filled-in prompt on the
Recruiting Dataset. The accepted resume is Resume~[2]. Sensitive
information has been redacted. For better display, Resumes~[1]-[4] and Example LLM output are heavily truncated.}
\label{tab:prompt_example}
\begin{tabular}{p{0.96\textwidth}}
\toprule
\textbf{Instruction/heuristics for person-job fit} (truncated) \\
Carefully verify that each candidate meets ALL of the following
mandatory criteria when explicitly stated in the job description: \\
\quad $\bullet$ Education: Required degree level and relevant major. \\
\quad $\bullet$ Certifications \& Licenses: Mandatory
professional qualifications (e.g., physician's license, CPA). \\
\quad $\bullet$ Technical Skills: ...\\
Critical Rule: The more explicitly stated mandatory criteria
a candidate fails to meet, the less matching they are to the job. \\
\midrule
\textbf{Job Description} (truncated) \\
title: Enterprise Account Manager/Director; job type: Full-Time;
location: England, UK; minimum degree: Master; required experience:
8+ years. \textit{Requirements:} Full UK driving license; strong
customer-facing skills; 8+ years of IT / computer-network sales; $\ldots$ \\
\midrule
\textbf{Resume [1]} -- Bachelor in Electrical \& Electronics; 4 yrs Technical Sales Engineer; 2 yrs Tech.
Sales Engineer; 1 yr Regional Business Manager ; $\ldots$ \\
\textbf{Resume [2]} \textbf{(accepted)} -- 2+ yrs External Account
Manager; 2 yrs Internal Account Manager; 1 yr Account
Manager; strong channel-distribution profile with
vendors including Cisco, HP, Dell, VMware ; $\ldots$ \\
\textbf{Resume [3]} -- 4 yrs Cisco Sales Specialist; 2 yrs
Internal Sales Executive; Apple retail / SME B2B; $\ldots$ \\
\textbf{Resume [4]} -- $\ldots$ \\
\midrule
\textbf{Example LLM Output} \\
\texttt{<think>The role emphasizes enterprise channel sales with 8+
years of IT / networking sales experience and a master's degree
preferred. Resume~[2] shows strong channel-distribution experience
with enterprise IT vendors$\ldots$ Resume~[1] is more technical
pre-sales oriented$\ldots$ Resume~[3] has Cisco-specific channel
experience but fewer years of enterprise sales$\ldots$</think>} \\
\texttt{<answer> [2] > [3] > [1] > [4] </answer>}\\
\bottomrule
\end{tabular}
\end{table*}

\section{Details on Dataset and Preprocessing}
\label{appendix:More Details on Dataset and Preprocessing}

\paragraph{Recruiting Dataset} The talent-job pairs are provided by a hiring solution company. The original resumes/job posts are parsed into text fields using techniques such as OCR. Some of the information is further corrected by humans. All sensitive information, such as names, contacts, college names, and company names, has been either removed or converted into numeric IDs. Example resume and job post are shown in \Cref{tbl:example_resume} and \Cref{tbl:example_job}, respectively.

\paragraph{AliYun Dataset} The 2019 Alibaba job-resume intelligent matching competition provided resume-job data that is already desensitized and parsed into a collection of text fields. There are 12 fields in a resume (\Cref{tbl:example_resume}) and 11 fields in a job post (\Cref{tbl:example_job}) used during training/validation/testing. Sensitive fields such as ``\chinese{居住城市}'' (living city) were already converted into numeric IDs. ``\chinese{工作经验}'' (work experience) was processed into a list of keywords. Overall, the average length of a resume or a job post in the AliYun dataset is much shorter than that of the Recruiting dataset (see \Cref{tbl:train_dset}).

We present the overall, training, and test dataset statistics in \Cref{tbl:train_dset} and \Cref{tbl:test_dset}.
\begin{table*}[h]
  \centering
    \caption{Example resume from the Recruiting dataset and AliYun dataset. The Recruiting dataset contains resumes in both English and Chinese, while the AliYun dataset contains resumes only in Chinese. All documents are available as a collection of text fields, and are formatted into a single string as shown above for \confitsimple{} training.}
  \label{tbl:example_resume}

  \scalebox{0.85}{
    \begin{tabular}{p{0.55\linewidth} p{0.45\linewidth}}
      \toprule
      \multicolumn{1}{c}{\textbf{$R$ from Recruiting Dataset}} &
      \multicolumn{1}{c}{\textbf{$R$ from AliYun Dataset}} \\
      \midrule
      \#\# current location
      & \#\# \chinese{居住城市}\\
      San Jose
      & 551 \\
      \#\# highest degree
      & \#\# \chinese{期望工作城市}\\
      BACHELOR
      & 551,-,-\\
      \#\# languages
      & \#\# \chinese{学历}\\
      THAI, ENGLISH, MANDARIN
      & \chinese{大专}\\
      \#\# top education
      & \#\# \chinese{期望工作行业}\\
      I graduated in [University] in Business Administration with BACHELOR degree in 2002. The college is in world top 200.
      & \chinese{房地产/建筑/建材/工程}\\
      \#\# most recent experience
      & \#\# \chinese{期望工作类型}\\
      I am currently working as a Pan America Sales Manager in [Company] for 19 years. ...
      & \chinese{工程造价/预结算}\\
      \#\# second experience
      & \#\# \chinese{当前工作行业}\\
      I am currently working as a Global Account Manager in [Company] for 20 years in Taiwan, China, and USA. ...
      & \chinese{房地产/建筑/建材/工程}\\
      \bottomrule
    \end{tabular}
  }
\end{table*}
\begin{table*}[h]
    \centering
    \caption{Example job post from the Recruiting dataset and AliYun dataset. The Recruiting dataset contains job posts in both English and Chinese, while the AliYun dataset contains job posts only in Chinese. All documents are available as a collection of text fields, and are formatted into a single string as shown above for \confitsimple{} training.}
    \label{tbl:example_job}

    \scalebox{0.85}{
      \begin{tabular}{p{0.55\linewidth} p{0.45\linewidth}}
        \toprule
        \multicolumn{1}{c}{\textbf{$J$ from Recruiting Dataset}} & 
        \multicolumn{1}{c}{\textbf{$J$ from AliYun Dataset}} \\
        \midrule
        \#\# title 
        & \#\# \chinese{工作名称}\\
        Net Processor Architecture Modeling Engineer 
        & \chinese{工程预算}\\
        \#\# job type 
        & \#\# \chinese{工作城市}\\
        Full-Time 
        & 719\\
        \#\# minimum degree 
        & \#\# \chinese{工作类型}\\
        MASTER 
        & \chinese{工程造价/预结算}\\
        \#\# required skills 
        & \#\# \chinese{招聘人数}\\
        RTOS, Neural Network, DSP, C, C++ 
        & 3\\
        \#\# preferred skills 
        & \#\# \chinese{薪资}\\
        Operations API, integration 
        & 1000-2000\chinese{元/月}\\
        \#\# required languages 
        & \#\# \chinese{是否要求出差}\\
        MANDARIN, ENGLISH 
        & \chinese{否}\\
        \#\# required experience 
        & \#\# \chinese{工作年限}\\
        more than 3 years 
        & 1-3\chinese{年}\\
        \#\# summaryText 
        & \#\# \chinese{最低学历}\\
        As firmware engineer at [Company], you will leverage your experience in protocol stack development, coding, microcontroller/microprocessor design and integration firmware for diverse product offering... 
        & \chinese{大专}\\
        \#\# requirementText 
        & \#\# \chinese{工作描述}\\
        Master's degree with 3+ years of experience in EE or CS or proof of skills in related fields, with practical engineering... 
        & \chinese{能够独立完成土建专业施工预算编制工作；熟练掌握工程造价相关软件...}\\
        \bottomrule
      \end{tabular}
    }
\end{table*}

\section{Details on Dataset Postprocessing}
\label{appendix:More Details on Dataset Postprocessing}

We describe how we convert raw binary-label data into the sliding-window format used for SFT distillation and RL training.
A ``window'' refers to a group of $k=4$ candidate resumes paired with a single job post.

\subsection{From Binary Labels to Ranking Windows}
\label{appendix:binary_to_window}

The raw datasets provide binary labels $y_i \in \{0, 1\}$ indicating whether resume $R_i$ was accepted by job $J$.
As discussed in Section~2, these labels are both sparse (less than 0.05\% of $(R, J)$ pairs are annotated) and noisy (many unlabelled resumes simply reflect that the candidate never applied).
Directly casting $(J, R_i, y_i)$ triples into a ranking task is therefore suboptimal.
 
Instead, we construct ranking windows on top of an existing
embedding-based retriever. Specifically, for every job post $J$ in the
training set, we first use \textsc{ConFit v2} to retrieve its top-20
candidate resumes $\mathcal{R}_{1:20}$ from the global resume pool.
We then partition $\mathcal{R}_{1:20}$ into positives
$\mathcal{P} = \{R \in \mathcal{R}_{1:20} : y_R = 1\}$ and
negatives $\mathcal{N} = \{R \in \mathcal{R}_{1:20} : y_R \ne 1\}$,
where resumes not present in the interaction label table are treated
as negatives. We then skip any job that satisfies one of the following
filters:
\begin{enumerate}
  \item $|\mathcal{R}_{1:20}| < 20$: not enough retrieved candidates
  to form a top-20 pool (typically due to extremely small-vertical
  job posts);
  \item $|\mathcal{P}| = 0$: no positive candidate in the top-20;
  \item $|\mathcal{P}| \ge m_{\max}$ with $m_{\max} = 11$: too many
  positives in the top-20, which often indicates either a very
  generic job post or a label-noise issue;
  \item $|\mathcal{N}| < 3$: not enough negatives to form a
  4-resume window.
\end{enumerate}

\subsection{Per-Positive Random Negative Sampling}
\label{appendix:sampling}
 
For each retained job $J$ and each positive $R^+ \in \mathcal{P}$, we draw $s=3$ negatives uniformly from $\mathcal{N}$ to form a 4-resume window $W = \{R^+, R^-_1, R^-_2, R^-_3\}$.
We repeat this $n_{\text{rep}}=3$ times per positive, yielding at most $3|\mathcal{P}|$ windows per job, and discard any window whose 4 IDs coincide with an earlier window for the same job.
 
Finally, to eliminate position bias,
we randomly shuffle the four resumes before filling them into the
prompt template.

\subsection{Difficulty Annotation}
\label{appendix:difficulty}
 
We annotate each window with an empirical per-sample
accuracy. Concretely, for each window $(J, W, \mathbf{y})$, we query
Qwen3-8B (under the same prompt template as Appendix~\ref{appendix:more_detail_on_rerank_prompt})
for $n = 5$ times with sampling temperature $0.6$, top-$p = 0.95$, and
top-$k = 20$, and record
the fraction of runs in which the accepted resume $R^+$ is correctly
placed at the top of the predicted ordering. We denote this quantity
$\bar{r} \in \{0.0, 0.2, 0.4, 0.6, 0.8, 1.0\}$. Windows with
$\bar{r} < 0.4$ are treated as ``hard'' (and potentially noisy), as mentioned in \Cref{subsec:RL Data Processing}.

\section{Details on Inference Algorithm}
\label{sec:details_on_inference_algorithm}

At inference time, LLM reranker is applied to the top-\(N\) resumes for each job by ConFit v2, where we fix $N=20$ for all our experiments. Limited by context length, we adopt a sliding-window reranking strategy. Specifically, starting from the bottom of the ranked list, we repeatedly select a window of size \(k\), ask the LLM to rerank the resumes within the window, and then move the window upward by a fixed stride \(s < k\). When the window reaches the top of the list, one full pass is completed. Since a single pass is often insufficient to fully propagate strong candidates to the top, we repeat the entire process for \(t\) iterations. The selection of $t=2$ is discussed in \Cref{subsec:Re-Ranking Inference Algorithms}. We also provide the pseudo code for the inference stage in \Cref{alg:rearank_inference}.

\begin{algorithm}[H]
\caption{Multi-pass inference algorithm for ReaRank}
\label{alg:rearank_inference}
\begin{algorithmic}[1]
\State \textbf{Input:} Job post \(j\)
\State \textbf{Input:} Top-20 ranked resumes \(R = [r_1, r_2, \dots, r_{20}]\)
\State \textbf{Input:} Window size \(k\), stride \(s < k\), number of iterations \(t\)
\State \textbf{Output:} Final reranked list \(R\)

\For{$iter = 1$ to $t$}
    \State $start \gets 20-k+1$
    \While{$start \geq 1$}
        \State Let \(W = R[start : start+k-1]\)
        \State Use the LLM to rerank resumes in \(W\) given job \(j\)
        \State Update \(R[start : start+k-1]\) using the reranked order
        \If{$start = 1$}
            \State \textbf{break}
        \EndIf
        \State $start \gets \max(1, start-s)$
    \EndWhile
\EndFor

\State \Return \(R\)
\end{algorithmic}
\end{algorithm}

In our experiments, we fix $k=4, s=2$ to strike a balance between performance and efficiency. We use \framework{} (without SFT) to do ablation studies on $k,s$. While the performance of $k=4, s=2$ is slightly worse than $k=3,s=1$, it yields only half comparisons for each iteration, which is important in practice. The complete results of ablating $k,s$ are shown in \Cref{tbl:ablation_ks}.

\begin{table}[H]
  \caption{Ablation study on window size $k$ and stride $s$}
  \label{tbl:ablation_ks}
  \centering
  \small
  \setlength{\tabcolsep}{6pt}
  \begin{tabular}{lcccc}
  \toprule
  Setting & nDCG@10 & Recall@10 & Average & Comp./iter. \\
  \midrule
  $k=2,s=1$ & 56.68 & 66.01 & 61.35 & 19 \\
  $k=3,s=1$ & \textbf{57.99} & \textbf{67.75} & \textbf{62.87} & \textbf{18} \\
  $k=3,s=2$ & 57.04 & 66.31 & 61.68 & 10 \\
  $k=4,s=1$ & 56.28 & 66.10 & 61.19 & 17 \\
  $k=4,s=2$ & \second{57.96} & \second{67.10} & \second{62.53} & \second{9} \\
  $k=4,s=3$ & 56.69 & 66.22 & 61.46 & 7 \\
  \bottomrule
  \end{tabular}
\end{table}

\section{Details on RL Implementation}
\label{appendix:details on RL implementation}

For RL training in \framework{}, we adopt Group Relative Policy Optimization (GRPO) \cite{shao2024deepseekmathpushinglimitsmathematical,deepseekr1} for training. Given an input \(x\), we sample a group of responses
\[
G = \{o_1, o_2, \dots, o_{|G|}\}
\sim \pi_\theta(\cdot \mid x).
\]
For each response \(o_i\), a rule-based reward \(R_i\) is computed. 
The group-relative advantage is then defined as
\[
\hat{A}_i = \frac{R_i - \mathrm{mean}(R)}{\mathrm{std}(R)},
\]
where the mean and standard deviation are computed over all rewards in the sampled group. Then, the training objective is
\[
\begin{aligned}
\mathcal{J}_{\mathrm{GRPO}}(\theta)
&=
-\frac{1}{|G|}
\sum_{i=1}^{|G|}
\frac{1}{|o_i|}
\sum_{t=1}^{|o_i|}
\Biggl[
\frac{
\pi_\theta(o_{i,t}\mid x, o_{i,<t})
}{
\bigl[\pi_\theta(o_{i,t}\mid x, o_{i,<t})\bigr]_{\mathrm{nograd}}
}
\hat{A}_i
\\
&\qquad\qquad
-\beta D_{\mathrm{KL}}
\!\left(
\pi_\theta(\cdot\mid x,o_{i,<t})
\;\|\;
\pi_{\mathrm{ref}}(\cdot\mid x,o_{i,<t})
\right)
\Biggr].
\end{aligned}
\]
Here, \(\pi_\theta\) is the current policy, \(\pi_{\mathrm{ref}}\) is a fixed reference policy, and \(\beta\) controls the strength of KL regularization. We use the same hyperparameter setting as ReaRank \citep{zhang2025rearankreasoningrerankingagent}, including \(\mathrm{learning\ rate} = 10^{-6}\), $|G|=32$, and $\beta=0.01$. We train $2$ epochs for all the experiments.

\begin{algorithm}[H]
\caption{RL training loop for ReaRank}
\label{alg:rearank_rl}
\begin{algorithmic}[1]
\State \textbf{Input:} Job-resume window \((j, W)\), where \(W=\{r_1,r_2,r_3,r_4\}\)
\State \textbf{Input:} Accepted resume index \(y \in \{1,2,3,4\}\)
\State \textbf{Input:} Policy model \(\pi_\theta\)

\For{each training step}
    \State Sample a mini-batch of \((j, W, y)\)
    \For{each instance}
        \State Compute \(\mathrm{nDCG}_{\text{old}} = \mathrm{nDCG}@4(W, y)\)
        \State Sample a reranked order \(\hat{W} \sim \pi_\theta(\cdot \mid j, W)\)
        \State Compute \(\mathrm{nDCG}_{\text{new}} = \mathrm{nDCG}@4(\hat{W}, y)\)
        \State Compute reward \(R\) according to \Cref{rearank_formula}
        \State Compute loss from \(R\)
    \EndFor
    \State Update \(\theta\) using the averaged mini-batch loss
\EndFor
\end{algorithmic}
\end{algorithm}

We list the pseudo code of our RL training loop for ReaRank in \Cref{alg:rearank_rl}. For the variant of Rank-R1 \citep{zhuang2025rankr1enhancingreasoningllmbased}, the model is expected to output the best-matching resume in the window, with binary reward $R=1$ if correct otherwise $R=0$. For re-ranking prompt template, please refer to \Cref{appendix:more_detail_on_rerank_prompt}.

\end{document}